\pdfoutput=1

\documentclass[11pt]{article}

\usepackage[]{ACL2023}
\usepackage{booktabs}
\usepackage{times}
\usepackage{latexsym}
\usepackage{graphicx} 
\usepackage{tcolorbox}
\usepackage{subcaption}
\usepackage[T1]{fontenc}

\usepackage[utf8]{inputenc}

\usepackage{microtype}
\usepackage{tabularx} 
\usepackage{inconsolata}
\usepackage{svg}
\usepackage{hyperref}

\title{%
DACP: Domain-Adaptive Continual Pre-Training of \\ 
Large Language Models for Phone Conversation Summarization\\
\texorpdfstring{%
  \raisebox{-0.2\height}{\includegraphics[height=7mm]{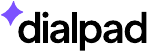}}%
}{}
}

\author{Xue-Yong Fu$^*$, Elena Khasanova$^*$, Md Tahmid Rahman Laskar$^*$, Harsh Saini$^*$, \\ \textbf{Shashi Bhushan TN} \\
\large{Dialpad Inc.} \\
  \texttt{\{xue-yong,elena.khasanova,tahmid.rahman,hsaini,sbhushan\}@dialpad.com}}

\begin{document}
\maketitle

\def\thefootnote{*}\footnotetext{\textbf{Equal Contributions. Sorted by the Last Name.}}\def\thefootnote{\arabic{footnote}}

\begin{abstract}


Large language models (LLMs) have achieved impressive performance in text summarization, yet their performance often falls short when applied to specialized domains 
that differ from their original pre-training distribution. While fine-tuning can improve summarization quality, it typically relies on costly and scarce high-quality labeled data. In this work, we explore continual pre-training as a scalable, self-supervised approach to adapt LLMs for downstream summarization tasks, particularly in the context of noisy real-world conversation transcripts. We conduct extensive experiments using large-scale, unlabeled business conversation data to investigate whether continual pre-training enhances model capabilities in conversational summarization. Our results demonstrate that continual pre-training yields substantial gains in both in-domain and out-of-domain summarization benchmarks, while maintaining strong generalization and robustness. We also analyze the effects of data selection strategies, 
providing practical guidelines for applying continual pre-training in summarization-focused industrial applications.

\end{abstract}
\section{Introduction}

LLMs have demonstrated remarkable performance in text summarization, even outperforming human-written summaries in various publicly available datasets \cite{pu2023summarizationdead,laskar2023systematicchatgpt}. 
This impressive capability of LLMs in generating high-quality summaries 
has led to the development of various LLM-powered summarization applications for practical use cases \cite{laskar-etal-2023-building}. 

However, real-world deployment of LLMs is associated with high inference costs \cite{wang2024comprehensive, lu2024small}. 
Therefore, smaller LLMs\footnote{We denote LLMs below 10B parameters as smaller LLMs.} are often preferred over their larger counterparts to reduce production costs \cite{fu-etal-2024-tiny}. 
Note that, despite the recent advances of LLMs in text summarization, recent research has found that the performance of LLMs, especially the cost-effective smaller ones, can drop sharply in downstream summarization tasks when the input differs from the initial data used during their pre-training \cite{afzal-etal-2024-adapteval}. 
Thus, it is important to adapt the smaller LLMs in the targeted domain before deploying them for real-world inference. 

Although smaller LLMs can be adapted to downstream tasks related to a certain domain by leveraging techniques like fine-tuning or instruction-tuning \cite{han2024parameter,zhang2023instruction}, this process requires the availability of human-annotated data, which can be challenging to obtain \cite{fu-etal-2024-tiny}. While this limitation can be addressed by leveraging larger closed-source LLMs for data annotation, their applicability in real-world scenarios is limited due to the privacy concerns of the customer data and the high cost of manually verifying LLM-annotated labels. In this regard, continual pre-training of smaller open-sourced LLMs on a vast amount of unlabeled internal data in a self-supervised fashion could be a potential solution for domain adaptation \cite{wu2024continual}.  

To this end, in this paper, we study the 
continual pre-training in the context of LLMs on real-world business conversational data. Our goal is to apply a data-centric solution and investigate whether they can help improve the performance in downstream summarization tasks related to real-world business conversations (e.g., meeting recaps, call summary and action items generation, etc.). 
Our extensive experiments demonstrate that 
continual pre-training \cite{wu2024continual} helps LLMs to improve their performance in downstream summarization tasks in the business conversational domain. 
Our major contributions in this paper are summarized below:

\begin{figure*}
    \centering
    \includegraphics[width=\linewidth]{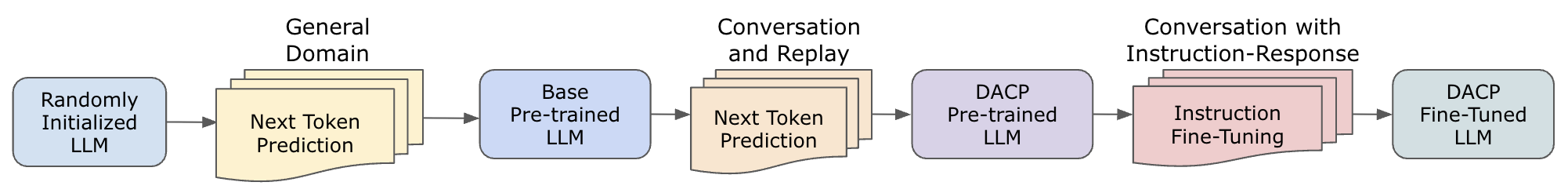}
    
    \caption{An overview of our proposed DACP framework of LLMs for business conversational tasks.}
    \label{fig:overview}
\end{figure*}

(i) We conduct extensive experiments to evaluate the effectiveness of self-supervised continual pre-training on large-scale unlabeled data for improving the performance of smaller LLMs in noisy, real-world business conversation summarization.

(ii) We present our data collection process for real-world business conversations 
and conduct extensive experiments to investigate how it impacts continual pre-training for domain adaptation.

(iii) We summarize key lessons from our experiments, offering practical guidelines for industry practitioners on when and how self-supervised continual pre-training can be effectively applied to business conversational summarization tasks.

\section{Related Work} 

Existing LLMs
are massively pre-trained on vast amounts of publicly available internet data using the self-supervised Next Token Prediction (NTP) objective \cite{gpt3,touvron2023llama,touvron2023llama2,openai2023gpt4,team2023gemini}. However, these public datasets 
can be significantly different than the proprietary data used in the real-world industrial scenario \cite{wu2023bloomberggpt}. As demonstrated by \citet{afzal-etal-2024-adapteval}, 
LLMs often underperform on real-world, domain-specific summarization compared to public benchmarks that reflect their pre-training data . 

To address this, continual pre-training via leveraging self-supervised learning on internal datasets could be useful to adapt existing LLMs to a specific domain \cite{wu2024continual}, as demonstrated by 
\cite{labrak2024biomistral,wu2024pmc,gururangan2020don}. Nonetheless, prior research on continual pre-training of LLMs is mostly limited to certain domains, such as biomedicine \cite{labrak2024biomistral,wu2024pmc,gururangan2020don} or finance \cite{xie2023efficient}. No prior research has studied the effectiveness of domain adaptation via continual pre-training on noisy conversational data. Since utilization of LLMs on conversational data is on the rise\footnote{\url{https://masterofcode.com/blog/llm-for-call-centers}} for real-world use cases \cite{laskar-etal-2023-building,nathan2024can}, it is important to investigate how to effectively utilize vast amounts of 
unlabeled ASR-generated conversation transcripts to successfully adapt LLMs to downstream tasks related to real-world business conversations. 

 In this paper, we aim to address the gap in the prior research. Our focus is to investigate the effectiveness of continual pre-training for domain adaptation by leveraging large amounts of unlabeled business conversations. Based on our extensive experiments, we provide our insights on (i) how we select the data for continual pre-training and why we choose a particular strategy, (ii) what pre-training strategy is followed and why, and (iii) how helpful continual pre-training is to adapt LLMs to various summarization tasks related to business conversations. These findings will help industries working with conversational data 
 to effectively utilize LLMs for real-world use cases. 




 


\section{Methodology}
\label{method}

An overview of our methodology is shown in Figure \ref{fig:overview}. Below, we describe the overall process. 



\begin{figure*}
    \centering
    \includegraphics[width=\linewidth]{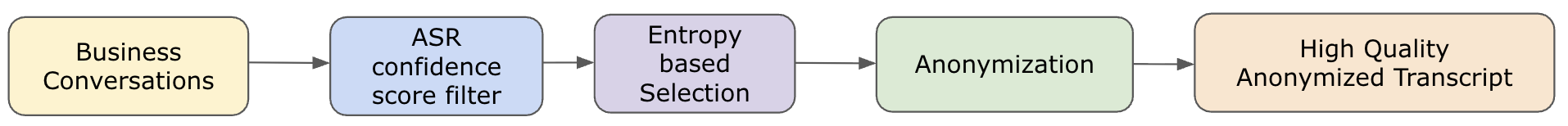}
 
    \caption{Our High-Quality Anonymized Transcript Data Selection Methodology for Pre-Training.}
    \label{fig:data_selection}
\end{figure*}
\subsection{Domain Adaptive Continual Pre-Training (DACP)} 
LLMs are initially pre-trained on large unlabeled text corpora with the self-supervised NTP objective \cite{zhao2023survey}. Since our focus is to leverage unlabeled business conversations, we also utilize self-supervised learning based on the NTP objective for continual pre-training. Nonetheless, this is still a data-hungry task that requires the data to be representative of the target domain and at the same time allowing the model to retain its general capabilities. Thus, we compose our dataset of two parts: real-world business conversational data collected from Dialpad\footnote{\url{https://www.dialpad.com/ca/}} and external experience replay data \cite{sun2020distill,chen2023meditron}, with a pre-decided maximum token budget of roughly 25B tokens for each part as described below.
\subsubsection{\textbf{In-domain Pre-Training Data}}
\label{sec:indomain_data}
Our internal dataset consists of English transcripts from real business conversations, generated via an in-house ASR system. 
To ensure 
diversity, we initially sample 50M 
transcripts from diverse organizations, having a minimum duration of 120s with at least two speakers. From these, we select 25M transcripts ($\approx$ 25B tokens) with the highest token type entropy scores, following \citet{xie2023efficient}. The data is anonymized using Google Cloud Data Loss Prevention\footnote{\url{https://cloud.google.com/security/products/}} with custom info types, as described in \cite{zhang2024data}. See Figure \ref{fig:data_selection} 
for an overview of our data 
construction methodology.

\subsubsection{\textbf{Experience Replay Data}}
One of the major challenges of continual pre-training 
is experiencing catastrophic forgetting \cite{sun2020distill}. A common mitigation strategy, known as experience replay \cite{rolnick2019experience}, involves incorporating data previously encountered during initial pre-training into the continual pre-training dataset \cite{sun2020distill,chen2023meditron}. Following the findings from  \citet{gu2024cmr}, we combined 25B replay tokens with 25B domain-specific tokens to construct a 50B continual pre-training dataset. The data for 25B replay tokens were randomly sampled from FineWeb-Edu \cite{penedo2024fineweb}.
\subsection{In-domain Instruction Fine-Tuning Data}
\label{sec:indomain_ft}
We collected some conversational data and curated instructions for various text generation and classification tasks related to conversations. 
To maintain the general instruction-following capabilities of the model, we also included general instructions that were generated using GPT-4 following the self-instruct methodology \cite{wang-etal-2023-self-instruct,openai2023gpt4}. 
GPT-4 was then used to generate responses for all of the selected instructions, which were subsequently evaluated and refined by human reviewers to create the in-domain instruction fine-tuning dataset containing 
84585 examples. 

\subsection{Downstream Summarization Tasks}

For evaluation, we select datasets from domain-specific internal benchmarks, as well as external public benchmarks. 
\paragraph{Internal Benchmarks:}
Our internal benchmarks consist of the following two tasks (the fine-tuning dataset also includes the training data of each of these tasks).

\textbf{(i) Action Items}: This task focuses on summarizing the list of actionable items from the conversation transcript. Each action item is a short description of an activity that should occur after the conversation has ended. This dataset consists of 120 instances. 

\definecolor{attachedColor}{HTML}{e0efff}
\definecolor{attachedColor2}{HTML}{f3f3f3}
\definecolor{attachedColor3}{HTML}{FFE5CC}
\definecolor{attachedColor4}{HTML}{FFCCCC}
\begin{tcolorbox}[
boxrule=0.25pt,   
  colback=attachedColor2,    
  colframe=black,           
  colbacktitle=attachedColor, 
  coltitle=black,           
  title={{Prompt: Action Items}},
  fonttitle=\bfseries,      
  fontupper=\small          
]

For the conversation given below, generate a newline-separated list of work, business, or service-related TODO tasks that should be completed after the conversation. Each task is a one-sentence summary of the action to be taken.\\

Transcript: [Call Conversation Transcript]

\end{tcolorbox}


 \textbf{(ii) Support Call Summarization}: The task is to generate a concise conversation summary. This task may also require the model to generate the summary in a specified length (long, medium, or short) or format (e.g. in bullet points). The dataset contains 204 instances. 

 \definecolor{attachedColor}{HTML}{e0efff}
\definecolor{attachedColor2}{HTML}{f3f3f3}
\definecolor{attachedColor3}{HTML}{FFE5CC}
\definecolor{attachedColor4}{HTML}{FFCCCC}
\begin{tcolorbox}[
boxrule=0.25pt,   
  colback=attachedColor2,    
  colframe=black,           
  colbacktitle=attachedColor, 
  coltitle=black,           
  title={{Prompt: Support Call Summarization}}, 
  fonttitle=\bfseries,      
  fontupper=\small          
]

Generate a \{Length Type\} summary of the following conversation \{Format\} without assessing its quality. \\\\
Transcript: [Call Conversation Transcript]

\end{tcolorbox}


\begin{table*}[t!]
\tiny
\centering
\setlength{\tabcolsep}{2pt}
\resizebox{\textwidth}{!}{
\begin{tabular}{l|ccccc|ccccc|ccccc|ccccc}
\hline
\textbf{Model} & \multicolumn{5}{c|}{\textbf{Action Items}} & \multicolumn{5}{c|}{\textbf{Support Call Summarization}} & \multicolumn{5}{c|}{\textbf{QMSUM}} & \multicolumn{5}{c}{\textbf{QMSUM-I}} \\ 
\cline{2-21}
& \textbf{R-1} & \textbf{R-2} & \textbf{R-L} & \textbf{A-S} & \textbf{B-S} 
& \textbf{R-1} & \textbf{R-2} & \textbf{R-L} & \textbf{A-S} & \textbf{B-S}  
& \textbf{R-1} & \textbf{R-2} & \textbf{R-L} & \textbf{A-S} & \textbf{B-S}  
& \textbf{R-1} & \textbf{R-2} & \textbf{R-L} & \textbf{A-S} & \textbf{B-S}  \\
\hline
\textbf{LLaMA-3.1-8B} 
& 56.31 & 36.07 & 43.24 & 35.56 & 71.65 
& 59.07 & 32.51 & 44.43 & 46.00 & 73.89
& 18.38 & 3.96 & 12.24 & 10.23 & 53.68
& 24.19 & 7.41 & 14.06 & 41.10 & 52.63 \\
\textbf{LLaMA-3.1-8B-DACP-50M} 
& 56.83 & 37.48 & 44.30 & 37.13 & 72.55 
& 59.39 & 32.38 & 44.12 & 48.45 & 74.03
& 23.61 & 5.28 & 15.40 & 10.82 & 55.68
& 35.20 & 12.53 & 20.76 & 52.26 & 60.99 \\
\hline
\textbf{Mistral-V0.3-7B} 
& 53.95 & 33.35 & 41.01 & 31.17 & 70.40 
& 56.71 & 29.14 & 41.31 & 45.37 & 72.48
& 8.79 & 2.01 & 6.01 & 15.28 & 48.08
& 11.47 & 3.44 & 6.70 & 55.41 & 40.92 \\
\textbf{Mistral-V0.3-7B-DACP-50M} 
& 57.36 & 36.66 & 43.40 & 34.72 & 72.57 
& 59.04 & 31.91 & 43.66 & 47.95 & 73.99
& 23.39 & 5.76 & 15.40 & 14.99 & 55.64
& 27.27 & 9.77 & 15.69 & 55.16 & 51.82 \\
\hline
\end{tabular}
}
\caption{\small{Performance comparison between DACP (internal + replay) fine-tuned and original fine-tuned LLaMA and Mistral models across internal business conversational tasks and external benchmarks (QMSUM, QMSUM-I). Here, `R' denotes `ROUGE' \cite{lin2004rouge}, `A-S' denotes `AlignScore' \cite{zha-etal-2023-alignscore}, and `B-S' denotes `BERTScore' \cite{zhang2019bertscore}.}}
\label{tab:main_results}
\end{table*}
\paragraph{External Benchmarks:}
Our external benchmark uses the publicly available QMSUM dataset \cite{zhong2021qmsum}, relevant to the internal business use cases (e.g., meeting summarization):

\textbf{(i) QMSUM}: We use the QMSUM dataset \cite{zhong2021qmsum} which requires the generation of a meeting summary based on the given query. This dataset contains 281 samples requiring the meeting summary for a given query. 

\textbf{(ii) QMSUM-I}: We use the instruction-focused version of QMSUM, the QMSUM-I dataset from \citet{fu-etal-2024-tiny}, which requires the generation of overall meeting summaries based on three types of instructions: \textit{Long}, \textit{Medium}, and \textit{Short}. This dataset consists of 111 test instances. 


We use the prompts constructed\footnote{We only use the single-query setup since the multi-query setup requires longer context \cite{laskar-etal-2024-query} but our models are pre-trained and fine-tuned on 8K context length.} by \citet{laskar-etal-2024-query} in these external datasets for evaluation. 

\subsection{Models} 
While there are numerous LLMs available currently, we select the base versions of the following two LLMs for our study: LLaMA-3.1-8B \cite{dubey2024llama3} and Mistral-v0.3-7B \cite{jiang2023mistral}. We select Mistral-v0.3-7B since it demonstrates better performance than other LLMs of the same size (7B parameters) on conversational datasets \cite{laskar-etal-2024-query}; and LLaMA-3.1-8B \cite{touvron2023llama}, due to its widespread adoption in real-world tasks \cite{meta2025llama_industry}. 
\subsection{Training and Evaluation Settings} 
We conduct experiments on a six-node cluster, each with 8 x NVIDIA A100 80GB GPUs. The implementation was done using Huggingface Transformers \cite{wolf2019huggingface} and DeepSpeed \cite{aminabadi2022deepspeed}. After small-scale experiments with different hyperparameters, we select the following values: the learning rate was set as 2e-6, the context length was 8000, and pre-training was conducted for a total of 1 epoch. 
The pre-trained model was then fine-tuned 
for 3 epochs 
and finally evaluated in terms of ROUGE \cite{lin2004rouge}, BERTScore \cite{zhang2019bertscore}, and AlignScore \cite{zha-etal-2023-alignscore} using the \textit{LLM Evaluate} \cite{saini-etal-2025-llm} tool. 


\section{Results and Discussion}
\subsection{Main Findings} 

In this section, we present our experimental results to investigate the effectiveness of DACP. We compare the models pre-trained using the DACP approach against the original base pre-trained LLMs. For this purpose, we fine-tune both the DACP and the base models on our in-domain instruction fine-tuning dataset (see Table \ref{tab:main_results} for the results).

\textbf{Performance on Internal Benchmarks:} 
We find that in text generation tasks (Action Items and Summarization), while DACP did not bring a huge gain in performance for LLaMA-3.1-8B, it led to a major performance boost for the smaller, Mistral-V0.3-7B, on both tasks. More specifically, it resulted in an increase of 6.32\% and 4.11\% on Action Items and Support Call Summarization, in terms of ROUGE-1, respectively. Interestingly, in terms of the AlignScore metric for factual consistency, we observe higher gains in performance for both models in comparison to textual similarity metrics (e.g., ROUGE and BERTScore). 

\textbf{Performance on External Benchmarks:}
We also observe the effectiveness of our proposed DACP approach on the external benchmarks, where the performance increases for both Mistral and LLaMA. More specifically, the average gains in performance are by 38.15\% and 9.75\% for LLaMA, and by 150.04\% and 20.74\% for Mistral, in terms of ROUGE-1 and BERTScore, respectively.  
This shows that our DACP approach helps the model generalize better across datasets and tasks that are not included in the fine-tuning dataset. 


\subsection{Ablation Study} 

\begin{figure}[t!]
    \centering
    \includegraphics[width=\linewidth]{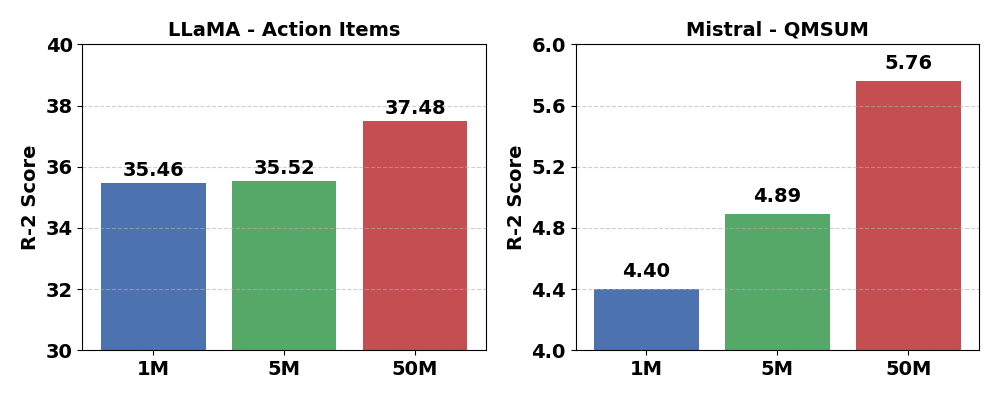}
    \caption{Ablation test results based on DACP data size: Action Items for LLaMA and QMSUM for Mistral.}
    \label{fig:ablation}
\end{figure}

To examine how the size of the DACP data affects model performance, we compare the performance of DACP models using 1M, 5M, and 50M examples (i.e., 1B, 5B, and 50B tokens, respectively) with the data mixture of 1:1: 50\% in-domain conversational data and 50\% replay data. Based on the results shown in Figure \ref{fig:ablation}, we find that more data is generally more useful for both models. 

\subsection{Qualitative Evaluation}
In our prior experiments, we observe in terms of automatic metrics that DACP helps improve the performance for both LLaMA and Mistral. In this section, we conduct a reference-free qualitative evaluation using an LLM Judge, the Gemini-2.5-Pro \cite{team2023gemini} model. The judge was prompted (see Appendix \ref{llm_judge_prompt} for the sample prompt) to select the better response output of the two model-generated responses (\textit{with} DACP vs \textit{without} DACP) in the internal datasets by considering factual correctness, adherence to instruction, and format following.  The task description and the input transcript were also provided as context for the LLM-judge. We find that on average, DACP wins 45\% of the time, in comparison to \textit{without} DACP (wins only 29\% of cases). 

\section{Conclusion and Future Work}
In this paper, we study how to effectively leverage vast amounts of unlabeled ASR-generated transcripts to adapt LLMs to handle real-world business conversational tasks. Based on extensive experiments, we 
observe that our proposed DACP technique helps LLMs to adapt effectively across downstream summarization tasks, demonstrating strong generalizability and robustness. This suggests that strategic 
data curation and processing, focusing on quality and diversity 
can 
lead to better model adaptation, a key consideration when dealing with large unlabeled 
industrial datasets.
In the future, we will explore the interplay between the model size and the data size in DACP-style training, alongside developing a new domain-specific benchmark with a broader task selection. 



\section*{Limitations} Note that our experiments are conducted 
on downstream summarization tasks only relevant to the target domain. 
Although extending experimentation to more domains, models, and tasks is prohibitively expensive due to the cost of computational resources, future work can focus on addressing these issues.  

\section*{Ethics Statement}

While using tools from various providers (e.g. Meta, Mistral AI, HuggingFace), we followed their licensing requirements accordingly. In terms of the models obtained through the training process described in the paper, they were used for research purposes only and so did not require safety evaluation. In this work, proprietary data containing sensitive information is used in the in-domain portion of the pretraining dataset as well as the instruction-following dataset described in sections \ref{sec:indomain_data} and \ref{sec:indomain_ft}, respectively. We protected the safety and privacy of the internal data used in the experiments by extensively anonymizing sensitive information with a robust method (see Appendix \ref{data_anonymization}). 
Following the privacy best practices \cite{narayanan2007breakanonymitynetflixprize}, we are not releasing these datasets to the public to completely eliminate the risk of sensitive data leakage.


\bibliography{anthology,custom}

\begin{thebibliography}{41}
\expandafter\ifx\csname natexlab\endcsname\relax\def\natexlab#1{#1}\fi

\bibitem[{Afzal et~al.(2024)Afzal, Chalumattu, Matthes, and Mascarell}]{afzal-etal-2024-adapteval}
Anum Afzal, Ribin Chalumattu, Florian Matthes, and Laura Mascarell. 2024.
\newblock \href {https://doi.org/10.18653/v1/2024.customnlp4u-1.8} {{A}dapt{E}val: Evaluating large language models on domain adaptation for text summarization}.
\newblock In \emph{Proceedings of the 1st Workshop on Customizable NLP: Progress and Challenges in Customizing NLP for a Domain, Application, Group, or Individual (CustomNLP4U)}, pages 76--85, Miami, Florida, USA. Association for Computational Linguistics.

\bibitem[{Aminabadi et~al.(2022)Aminabadi, Rajbhandari, Awan, Li, Li, Zheng, Ruwase, Smith, Zhang, Rasley et~al.}]{aminabadi2022deepspeed}
Reza~Yazdani Aminabadi, Samyam Rajbhandari, Ammar~Ahmad Awan, Cheng Li, Du~Li, Elton Zheng, Olatunji Ruwase, Shaden Smith, Minjia Zhang, Jeff Rasley, et~al. 2022.
\newblock Deepspeed-inference: enabling efficient inference of transformer models at unprecedented scale.
\newblock In \emph{SC22: International Conference for High Performance Computing, Networking, Storage and Analysis}, pages 1--15. IEEE.

\bibitem[{Brown et~al.(2020)Brown, Mann, Ryder, Subbiah, Kaplan, Dhariwal, Neelakantan, Shyam, Sastry, Askell et~al.}]{gpt3}
Tom Brown, Benjamin Mann, Nick Ryder, Melanie Subbiah, Jared~D Kaplan, Prafulla Dhariwal, Arvind Neelakantan, Pranav Shyam, Girish Sastry, Amanda Askell, et~al. 2020.
\newblock Language models are few-shot learners.
\newblock \emph{Advances in neural information processing systems}, 33:1877--1901.

\bibitem[{Chen et~al.(2023)Chen, Cano, Romanou, Bonnet, Matoba, Salvi, Pagliardini, Fan, K{\"o}pf, Mohtashami et~al.}]{chen2023meditron}
Zeming Chen, Alejandro~Hern{\'a}ndez Cano, Angelika Romanou, Antoine Bonnet, Kyle Matoba, Francesco Salvi, Matteo Pagliardini, Simin Fan, Andreas K{\"o}pf, Amirkeivan Mohtashami, et~al. 2023.
\newblock Meditron-70b: Scaling medical pretraining for large language models.
\newblock \emph{arXiv preprint arXiv:2311.16079}.

\bibitem[{Dubey et~al.(2024)Dubey, Jauhri, Pandey, Kadian, Al-Dahle, Letman, Mathur, Schelten, Yang, Fan et~al.}]{dubey2024llama3}
Abhimanyu Dubey, Abhinav Jauhri, Abhinav Pandey, Abhishek Kadian, Ahmad Al-Dahle, Aiesha Letman, Akhil Mathur, Alan Schelten, Amy Yang, Angela Fan, et~al. 2024.
\newblock The llama 3 herd of models.
\newblock \emph{arXiv preprint arXiv:2407.21783}.

\bibitem[{Fu et~al.(2024)Fu, Laskar, Khasanova, Chen, and Tn}]{fu-etal-2024-tiny}
Xue-Yong Fu, Md~Tahmid~Rahman Laskar, Elena Khasanova, Cheng Chen, and Shashi Tn. 2024.
\newblock \href {https://aclanthology.org/2024.naacl-industry.33} {Tiny titans: Can smaller large language models punch above their weight in the real world for meeting summarization?}
\newblock In \emph{Proceedings of the 2024 Conference of the North American Chapter of the Association for Computational Linguistics: Human Language Technologies (Volume 6: Industry Track)}, pages 387--394, Mexico City, Mexico. Association for Computational Linguistics.

\bibitem[{Gu et~al.(2024)Gu, Yang, Ding, Zhao, and Tan}]{gu2024cmr}
Jiawei Gu, Zacc Yang, Chuanghao Ding, Rui Zhao, and Fei Tan. 2024.
\newblock Cmr scaling law: Predicting critical mixture ratios for continual pre-training of language models.
\newblock \emph{arXiv preprint arXiv:2407.17467}.

\bibitem[{Gururangan et~al.(2020)Gururangan, Marasovi{\'c}, Swayamdipta, Lo, Beltagy, Downey, and Smith}]{gururangan2020don}
Suchin Gururangan, Ana Marasovi{\'c}, Swabha Swayamdipta, Kyle Lo, Iz~Beltagy, Doug Downey, and Noah~A Smith. 2020.
\newblock Don’t stop pretraining: Adapt language models to domains and tasks.
\newblock In \emph{Proceedings of the 58th Annual Meeting of the Association for Computational Linguistics}, pages 8342--8360.

\bibitem[{Han et~al.(2024)Han, Gao, Liu, Zhang, and Zhang}]{han2024parameter}
Zeyu Han, Chao Gao, Jinyang Liu, Jeff Zhang, and Sai~Qian Zhang. 2024.
\newblock Parameter-efficient fine-tuning for large models: A comprehensive survey.
\newblock \emph{arXiv preprint arXiv:2403.14608}.

\bibitem[{Jiang et~al.(2023)Jiang, Sablayrolles, Mensch, Bamford, Chaplot, Casas, Bressand, Lengyel, Lample, Saulnier et~al.}]{jiang2023mistral}
Albert~Q Jiang, Alexandre Sablayrolles, Arthur Mensch, Chris Bamford, Devendra~Singh Chaplot, Diego de~las Casas, Florian Bressand, Gianna Lengyel, Guillaume Lample, Lucile Saulnier, et~al. 2023.
\newblock Mistral 7b.
\newblock \emph{arXiv preprint arXiv:2310.06825}.

\bibitem[{Labrak et~al.(2024)Labrak, Bazoge, Morin, Gourraud, Rouvier, and Dufour}]{labrak2024biomistral}
Yanis Labrak, Adrien Bazoge, Emmanuel Morin, Pierre-Antoine Gourraud, Mickael Rouvier, and Richard Dufour. 2024.
\newblock Biomistral: A collection of open-source pretrained large language models for medical domains.
\newblock \emph{arXiv preprint arXiv:2402.10373}.

\bibitem[{Laskar et~al.(2023{\natexlab{a}})Laskar, Bari, Rahman, Bhuiyan, Joty, and Huang}]{laskar2023systematicchatgpt}
Md~Tahmid~Rahman Laskar, M~Saiful Bari, Mizanur Rahman, Md~Amran~Hossen Bhuiyan, Shafiq Joty, and Jimmy Huang. 2023{\natexlab{a}}.
\newblock \href {https://aclanthology.org/2023.findings-acl.29} {A systematic study and comprehensive evaluation of {C}hat{GPT} on benchmark datasets}.
\newblock In \emph{Findings of the Association for Computational Linguistics: ACL 2023}, pages 431--469, Toronto, Canada. Association for Computational Linguistics.

\bibitem[{Laskar et~al.(2023{\natexlab{b}})Laskar, Fu, Chen, and Bhushan~TN}]{laskar-etal-2023-building}
Md~Tahmid~Rahman Laskar, Xue-Yong Fu, Cheng Chen, and Shashi Bhushan~TN. 2023{\natexlab{b}}.
\newblock \href {https://doi.org/10.18653/v1/2023.emnlp-industry.33} {Building real-world meeting summarization systems using large language models: A practical perspective}.
\newblock In \emph{Proceedings of the 2023 Conference on Empirical Methods in Natural Language Processing: Industry Track}, pages 343--352, Singapore. Association for Computational Linguistics.

\bibitem[{Laskar et~al.(2024)Laskar, Khasanova, Fu, Chen, and Tn}]{laskar-etal-2024-query}
Md~Tahmid~Rahman Laskar, Elena Khasanova, Xue-Yong Fu, Cheng Chen, and Shashi~Bhushan Tn. 2024.
\newblock \href {https://doi.org/10.18653/v1/2024.emnlp-industry.86} {Query-{OPT}: Optimizing inference of large language models via multi-query instructions in meeting summarization}.
\newblock In \emph{Proceedings of the 2024 Conference on Empirical Methods in Natural Language Processing: Industry Track}, pages 1140--1151, Miami, Florida, US. Association for Computational Linguistics.

\bibitem[{Lin(2004)}]{lin2004rouge}
Chin-Yew Lin. 2004.
\newblock Rouge: A package for automatic evaluation of summaries.
\newblock In \emph{Text summarization branches out}, pages 74--81.

\bibitem[{Lu et~al.(2024)Lu, Li, Cai, Yi, Liu, Zhang, Lane, and Xu}]{lu2024small}
Zhenyan Lu, Xiang Li, Dongqi Cai, Rongjie Yi, Fangming Liu, Xiwen Zhang, Nicholas~D Lane, and Mengwei Xu. 2024.
\newblock Small language models: Survey, measurements, and insights.
\newblock \emph{arXiv preprint arXiv:2409.15790}.

\bibitem[{{Meta}(2025)}]{meta2025llama_industry}
{Meta}. 2025.
\newblock How organizations are using llama to solve industry challenges.
\newblock https://about.fb.com/news/2025/01/organizations-using-llama-solve-industry-challenges/.
\newblock Accessed: 2025-08-14.

\bibitem[{Narayanan and Shmatikov(2007)}]{narayanan2007breakanonymitynetflixprize}
Arvind Narayanan and Vitaly Shmatikov. 2007.
\newblock \href {http://arxiv.org/abs/cs/0610105} {How to break anonymity of the netflix prize dataset}.

\bibitem[{Nathan et~al.(2024)Nathan, Kumar, and Ingle}]{nathan2024can}
Varun Nathan, Ayush Kumar, and Digvijay Ingle. 2024.
\newblock Can probing classifiers reveal the learning by contact center large language models?: No, it doesn’t!
\newblock In \emph{Proceedings of the Fifth Workshop on Insights from Negative Results in NLP}, pages 92--100.

\bibitem[{OpenAI(2023)}]{openai2023gpt4}
OpenAI. 2023.
\newblock \href {http://arxiv.org/abs/2303.08774} {Gpt-4 technical report}.

\bibitem[{Penedo et~al.(2024)Penedo, Kydl{\'\i}{\v{c}}ek, Lozhkov, Mitchell, Raffel, Von~Werra, Wolf et~al.}]{penedo2024fineweb}
Guilherme Penedo, Hynek Kydl{\'\i}{\v{c}}ek, Anton Lozhkov, Margaret Mitchell, Colin Raffel, Leandro Von~Werra, Thomas Wolf, et~al. 2024.
\newblock The fineweb datasets: Decanting the web for the finest text data at scale.
\newblock \emph{arXiv preprint arXiv:2406.17557}.

\bibitem[{Pu et~al.(2023)Pu, Gao, and Wan}]{pu2023summarizationdead}
Xiao Pu, Mingqi Gao, and Xiaojun Wan. 2023.
\newblock Summarization is (almost) dead.
\newblock \emph{arXiv preprint arXiv:2309.09558}.

\bibitem[{Rolnick et~al.(2019)Rolnick, Ahuja, Schwarz, Lillicrap, and Wayne}]{rolnick2019experience}
David Rolnick, Arun Ahuja, Jonathan Schwarz, Timothy Lillicrap, and Gregory Wayne. 2019.
\newblock Experience replay for continual learning.
\newblock \emph{Advances in neural information processing systems}, 32.

\bibitem[{Saini et~al.(2025)Saini, Laskar, Chen, Mohammadi, and Rossouw}]{saini-etal-2025-llm}
Harsh Saini, Md~Tahmid~Rahman Laskar, Cheng Chen, Elham Mohammadi, and David Rossouw. 2025.
\newblock \href {https://aclanthology.org/2025.coling-industry.24/} {{LLM} evaluate: An industry-focused evaluation tool for large language models}.
\newblock In \emph{Proceedings of the 31st International Conference on Computational Linguistics: Industry Track}, pages 286--294, Abu Dhabi, UAE. Association for Computational Linguistics.

\bibitem[{Sun et~al.(2020)Sun, Wang, Zhang, and Zong}]{sun2020distill}
Jingyuan Sun, Shaonan Wang, Jiajun Zhang, and Chengqing Zong. 2020.
\newblock Distill and replay for continual language learning.
\newblock In \emph{Proceedings of the 28th international conference on computational linguistics}, pages 3569--3579.

\bibitem[{Team et~al.(2023)Team, Anil, Borgeaud, Wu, Alayrac, Yu, Soricut, Schalkwyk, Dai, Hauth et~al.}]{team2023gemini}
Gemini Team, Rohan Anil, Sebastian Borgeaud, Yonghui Wu, Jean-Baptiste Alayrac, Jiahui Yu, Radu Soricut, Johan Schalkwyk, Andrew~M Dai, Anja Hauth, et~al. 2023.
\newblock Gemini: a family of highly capable multimodal models.
\newblock \emph{arXiv preprint arXiv:2312.11805}.

\bibitem[{Touvron et~al.(2023{\natexlab{a}})Touvron, Lavril, Izacard, Martinet, Lachaux, Lacroix, Rozi{\`e}re, Goyal, Hambro, Azhar et~al.}]{touvron2023llama}
Hugo Touvron, Thibaut Lavril, Gautier Izacard, Xavier Martinet, Marie-Anne Lachaux, Timoth{\'e}e Lacroix, Baptiste Rozi{\`e}re, Naman Goyal, Eric Hambro, Faisal Azhar, et~al. 2023{\natexlab{a}}.
\newblock Llama: Open and efficient foundation language models.
\newblock \emph{arXiv preprint arXiv:2302.13971}.

\bibitem[{Touvron et~al.(2023{\natexlab{b}})Touvron, Martin, Stone, Albert, Almahairi, Babaei, Bashlykov, Batra, Bhargava, Bhosale et~al.}]{touvron2023llama2}
Hugo Touvron, Louis Martin, Kevin Stone, Peter Albert, Amjad Almahairi, Yasmine Babaei, Nikolay Bashlykov, Soumya Batra, Prajjwal Bhargava, Shruti Bhosale, et~al. 2023{\natexlab{b}}.
\newblock Llama 2: Open foundation and fine-tuned chat models.
\newblock \emph{arXiv preprint arXiv:2307.09288}.

\bibitem[{Wang et~al.(2024)Wang, Zhang, Zhang, Wu, Mo, Lu, Wang, Li, Xu, Tang et~al.}]{wang2024comprehensive}
Fali Wang, Zhiwei Zhang, Xianren Zhang, Zongyu Wu, Tzuhao Mo, Qiuhao Lu, Wanjing Wang, Rui Li, Junjie Xu, Xianfeng Tang, et~al. 2024.
\newblock A comprehensive survey of small language models in the era of large language models: Techniques, enhancements, applications, collaboration with llms, and trustworthiness.
\newblock \emph{arXiv preprint arXiv:2411.03350}.

\bibitem[{Wang et~al.(2023)Wang, Kordi, Mishra, Liu, Smith, Khashabi, and Hajishirzi}]{wang-etal-2023-self-instruct}
Yizhong Wang, Yeganeh Kordi, Swaroop Mishra, Alisa Liu, Noah~A. Smith, Daniel Khashabi, and Hannaneh Hajishirzi. 2023.
\newblock \href {https://doi.org/10.18653/v1/2023.acl-long.754} {Self-instruct: Aligning language models with self-generated instructions}.
\newblock In \emph{Proceedings of the 61st Annual Meeting of the Association for Computational Linguistics (Volume 1: Long Papers)}, pages 13484--13508, Toronto, Canada. Association for Computational Linguistics.

\bibitem[{Wolf et~al.(2020)Wolf, Debut, Sanh, Chaumond, Delangue, Moi, Cistac, Rault, Louf, Funtowicz et~al.}]{wolf2019huggingface}
Thomas Wolf, Lysandre Debut, Victor Sanh, Julien Chaumond, Clement Delangue, Anthony Moi, Pierric Cistac, Tim Rault, R{\'e}mi Louf, Morgan Funtowicz, et~al. 2020.
\newblock Transformers: State-of-the-art natural language processing.
\newblock In \emph{Proceedings of the 2020 conference on empirical methods in natural language processing: system demonstrations}, pages 38--45.

\bibitem[{Wu et~al.(2024{\natexlab{a}})Wu, Lin, Zhang, Zhang, Xie, and Wang}]{wu2024pmc}
Chaoyi Wu, Weixiong Lin, Xiaoman Zhang, Ya~Zhang, Weidi Xie, and Yanfeng Wang. 2024{\natexlab{a}}.
\newblock Pmc-llama: toward building open-source language models for medicine.
\newblock \emph{Journal of the American Medical Informatics Association}, page ocae045.

\bibitem[{Wu et~al.(2023)Wu, Irsoy, Lu, Dabravolski, Dredze, Gehrmann, Kambadur, Rosenberg, and Mann}]{wu2023bloomberggpt}
Shijie Wu, Ozan Irsoy, Steven Lu, Vadim Dabravolski, Mark Dredze, Sebastian Gehrmann, Prabhanjan Kambadur, David Rosenberg, and Gideon Mann. 2023.
\newblock Bloomberggpt: A large language model for finance.
\newblock \emph{arXiv preprint arXiv:2303.17564}.

\bibitem[{Wu et~al.(2024{\natexlab{b}})Wu, Luo, Li, Pan, Vu, and Haffari}]{wu2024continual}
Tongtong Wu, Linhao Luo, Yuan-Fang Li, Shirui Pan, Thuy-Trang Vu, and Gholamreza Haffari. 2024{\natexlab{b}}.
\newblock Continual learning for large language models: A survey.
\newblock \emph{arXiv preprint arXiv:2402.01364}.

\bibitem[{Xie et~al.(2023)Xie, Aggarwal, and Ahmad}]{xie2023efficient}
Yong Xie, Karan Aggarwal, and Aitzaz Ahmad. 2023.
\newblock Efficient continual pre-training for building domain specific large language models.
\newblock \emph{arXiv preprint arXiv:2311.08545}.

\bibitem[{Zha et~al.(2023)Zha, Yang, Li, and Hu}]{zha-etal-2023-alignscore}
Yuheng Zha, Yichi Yang, Ruichen Li, and Zhiting Hu. 2023.
\newblock \href {https://aclanthology.org/2023.acl-long.634} {{A}lign{S}core: Evaluating factual consistency with a unified alignment function}.
\newblock In \emph{Proceedings of the 61st Annual Meeting of the Association for Computational Linguistics (Volume 1: Long Papers)}, pages 11328--11348, Toronto, Canada. Association for Computational Linguistics.

\bibitem[{Zhang et~al.(2024)Zhang, Paling, Preston~Thomas, Azizi, Gardiner, Humphreys, and Mailhot}]{zhang2024data}
Nathan Zhang, Anne Paling, Tania~Habib Preston~Thomas, Mahsa Azizi, Shayna Gardiner, Kevin Humphreys, and Frederic Mailhot. 2024.
\newblock Data anonymization for privacy-preserving large language model fine-tuning on call transcripts.
\newblock In \emph{Workshop on Computational Approaches to Language Data Pseudonymization (CALD-pseudo)}, page~64.

\bibitem[{Zhang et~al.(2023)Zhang, Dong, Li, Zhang, Sun, Wang, Li, Hu et~al.}]{zhang2023instruction}
Shengyu Zhang, Linfeng Dong, Xiaoya Li, Sen Zhang, Xiaofei Sun, Shuhe Wang, Jiwei Li, Runyi Hu, et~al. 2023.
\newblock Instruction tuning for large language models: A survey.
\newblock \emph{arXiv preprint arXiv:2308.10792}.

\bibitem[{Zhang et~al.(2019)Zhang, Kishore, Wu, Weinberger, and Artzi}]{zhang2019bertscore}
Tianyi Zhang, Varsha Kishore, Felix Wu, Kilian~Q Weinberger, and Yoav Artzi. 2019.
\newblock Bertscore: Evaluating text generation with bert.
\newblock In \emph{International Conference on Learning Representations}.

\bibitem[{Zhao et~al.(2023)Zhao, Zhou, Li, Tang, Wang, Hou, Min, Zhang, Zhang, Dong et~al.}]{zhao2023survey}
Wayne~Xin Zhao, Kun Zhou, Junyi Li, Tianyi Tang, Xiaolei Wang, Yupeng Hou, Yingqian Min, Beichen Zhang, Junjie Zhang, Zican Dong, et~al. 2023.
\newblock A survey of large language models.
\newblock \emph{arXiv preprint arXiv:2303.18223}.

\bibitem[{Zhong et~al.(2021)Zhong, Yin, Yu, Zaidi, Mutuma, Jha, Hassan, Celikyilmaz, Liu, Qiu et~al.}]{zhong2021qmsum}
Ming Zhong, Da~Yin, Tao Yu, Ahmad Zaidi, Mutethia Mutuma, Rahul Jha, Ahmed Hassan, Asli Celikyilmaz, Yang Liu, Xipeng Qiu, et~al. 2021.
\newblock Qmsum: A new benchmark for query-based multi-domain meeting summarization.
\newblock In \emph{Proceedings of the 2021 Conference of the North American Chapter of the Association for Computational Linguistics: Human Language Technologies}, pages 5905--5921.

\end{thebibliography}
\bibliographystyle{acl_natbib}

\appendix

\section{Appendix}
\label{appendix}

\subsection{Data Anonymization Details}
\label{data_anonymization}

We anonymize the sampled data using Google Cloud Data Loss Prevention (\url{https://cloud.google.com/security/products/}) service with custom info types following the approach described in \citet{zhang2024data}. We use a combination of masking tokens (e.g. <PERSON\_NAME\_1> instead of the real name) and noising tokens with custom replacements (e.g. replacing sensitive names with different gender-neutral names) to allow the model to learn the properties of sensitive tokens without exposing these tokens. To increase the transcript format diversity, we utilize variable speaker tags (e.g. speaker 1, name, initials, agent, customer, etc.) and randomly modify the transcripts to include timestamps, different spacing configurations between the turns, merging subsequent turns from the same speaker,

\subsection{Prompt for LLM Judge}
\label{llm_judge_prompt}

 \definecolor{attachedColor}{HTML}{e0efff}
\definecolor{attachedColor2}{HTML}{f3f3f3}
\definecolor{attachedColor3}{HTML}{FFE5CC}
\definecolor{attachedColor4}{HTML}{FFCCCC}
\begin{tcolorbox}[
boxrule=0.25pt,   
  colback=attachedColor3,    
  colframe=black,           
  colbacktitle=attachedColor, 
  coltitle=black,           
  title={Sample Prompt},
  fonttitle=\bfseries,      
  fontupper=\small          
]

You are provided with a task description, a transcript, and two responses generated by AI models (Model A and Model B). \\

Your goal is to evaluate the quality of each response based on the provided context.\\

Please rate each model on a Likert scale from 1 to 5 based on the criteria given below.\\

*Evaluation Criteria*\\

1: Factual Correctness: How accurately does the response reflect the information present in the transcript? Does it contain any information that is incorrect or not mentioned in the source?\\

2: Instruction Following: How well does the response adhere to all instructions and constraints outlined in the task description?\\

3: Clarity and Conciseness: Is the response easy to read, succinct, and to the point, avoiding unnecessary jargon, repetition, or filler words?\\

4: Structure and Formatting: Is the response use formatting appropriately for the task based on the requirement?\\

*Rating Scale*\\

1: The response is extremely poor.\\

2: The response is poor.\\

3: The response is average.\\

4: The response is good.\\

5: The response is excellent.\\

Please provide your complete evaluation in an Array of JSON objects format that contains the following keys: (i) ratings, and (ii) rationale. Here, ratings will contain an integer value between 1-5 (inclusive), while rationale will contain a brief justification for the rating. \\

The task description, transcript, and the responses generated by the AI models are given below. \\

[Task description (Action Items or Summarization)] \\

[Transcript]\\

[Model A Response]\\

[Model B Response]\\

\end{tcolorbox}

\end{document}